\documentclass[10pt,twocolumn,letterpaper]{article}

\usepackage[margin=0.75in]{geometry}
\usepackage[T1]{fontenc}
\usepackage{newtxtext,newtxmath}
\usepackage{booktabs}
\usepackage{graphicx}
\usepackage{amsmath}
\usepackage{microtype}
\usepackage{balance}
\usepackage{xcolor}
\usepackage{caption}
\usepackage{tikz}
\usetikzlibrary{positioning,arrows.meta}
\usepackage[colorlinks=true,linkcolor=blue!55!black,citecolor=blue!55!black,urlcolor=blue!55!black]{hyperref}

\hypersetup{
  pdftitle={Where Does Personalization Reside in Photo Editing? Evidence from One Photographer's Lightroom History},
  pdfauthor={Dongwu Qian},
  pdfsubject={Personalized photo editing; batch-shared evaluation}
}

\newcommand{\ev}{\ensuremath{\,\mathrm{EV}}}

\title{\vspace{-1.4em}\textbf{Where Does Personalization Reside in Photo Editing?\\[2pt]
Evidence from One Photographer's Lightroom History}}

\author{Dongwu Qian\\[2pt]
Independent Researcher\\[2pt]
\texttt{qiandongwu123@qq.com}}
\date{November 21, 2026}

\begin{document}
\maketitle
\thispagestyle{empty}

\begin{abstract}
\noindent
Personalized photo-editing systems must infer both a photographer's recurring preferences and the intent of a particular shoot, yet these signals are rarely evaluated separately. We study interpretable Lightroom adjustments in a longitudinal, private dataset of 3,193 edited photographs from 41 catalogs by one professional photographer, with holdouts formed at the batch level. We compare absolute and prior-relative targets, image-conditioned predictors, and estimators that receive a small set of edits from the target batch. Residual prediction improves validation error, but its advantage on held-out batches is small and not significant under a batch-clustered bootstrap (0.592 vs.\ 0.582\ev{} exposure MAE). In a batch-shared protocol, closed-form shrinkage using 4, 8, or 16 same-batch anchors reaches 0.466, 0.457, or 0.453\ev{} with the best estimator at each $k$, compared with 0.586--0.592\ev{} for the prior-only baseline; the paired reductions are significant under a batch-clustered bootstrap. Learned anchor-conditioned heads do not improve on this closed-form estimator, and a negative-trained prior already yields a low cold-image false-edit rate, leaving little benefit for an additional gate. The recommended estimator is closed-form and auditable, and it requires only a handful of same-batch edits. A render-space check against 383 native Lightroom exports reproduces the ranking: at $k=8$, online EB lowers $\Delta E_{00}$ against the prior by 0.41 ($p=0.042$), and the photographer's true parameters lower it by 0.78 ($p=0.034$). These results motivate evaluating personalization at the shoot level and treating parameter fidelity, cold-start risk, and user preference as distinct outcomes. An appendix extends the action space from six to thirteen global controls: a same-batch median look lowers rendered chroma toward the export, while a per-photograph residual head again fails to transfer. The study is limited to one photographer and does not claim that slider accuracy measures aesthetic preference.
\end{abstract}

\section{Introduction}

A photographer who edits hundreds of frames from one shoot does not treat each frame independently. After a few corrections, the rest of the shoot tends to receive a similar exposure lift, contrast curve, and color treatment. Commercial tools advertise that they can ``learn your style'' from an Adobe Lightroom\footnote{Adobe and Lightroom are trademarks or registered trademarks of Adobe Inc. This work is independent and not affiliated with or endorsed by Adobe.} catalog~\cite{lightroom}, but the underlying question is rarely decomposed: how much of a predicted adjustment comes from the image itself, how much from the photographer's long-run history, and how much from the shoot currently being edited?

This paper measures that decomposition on a real editing history. We instrument a private Adobe Lightroom catalog collection from one professional photographer: 41 catalogs, 3,193 edited photographs, six interpretable global adjustments (exposure, contrast, saturation, vibrance, clarity, texture), and batch-level holdouts that mimic the operational workflow ``edit a few frames, then apply the result to the rest of the shoot.'' The setting is deliberately narrow. The main experiments use six global adjustments; Appendix~\ref{app:e8} extends the action space to highlights, shadows, whites, blacks, eight-band HSL saturation, and dehaze, and shows that a same-batch median look transfers while a per-photograph residual head does not. We do not model local masks or the complete Adobe Lightroom rendering pipeline, and we do not claim that slider accuracy equals aesthetic preference. The narrowness is what makes the signals separable.

Our results support four statements. First, batch identity is a large source of label variation: a one-way decomposition attributes 36\% of exposure variance and 27\% of contrast variance to the batch, and batch medians deviate from the global median by $+0.84$ to $-0.62$\ev{}. A per-image predictor that ignores the shoot is therefore evaluated on the wrong unit.

Second, prior-relative (residual) targets improve in-domain validation error but do not transfer across shoots. On validation, the residual head reaches 0.171\ev{} against 0.195\ev{} for the absolute head. On 1,448 held-out photographs from 11 unseen batches the ordering reverses (0.592 vs.\ 0.582\ev{}), and a batch-clustered bootstrap gives a paired difference of $+0.010$\ev{} with a 95\% interval of $[-0.016,+0.033]$ ($p=0.51$). The earlier photograph-level test ($p=0.027$) overstated the evidence because it treated correlated photographs from the same shoot as independent.

Third, a small number of same-batch anchors is the robust signal. Under a batch-shared protocol---one anchor set per batch, shared by all targets in that batch---closed-form shrinkage reduces exposure MAE from 0.586--0.592\ev{} to 0.466\ev{} at $k=4$, 0.457\ev{} at $k=8$, and 0.453\ev{} at $k=16$ (best arm at each $k$). The paired reductions against the prior are 0.11--0.14\ev{} and remain significant under a batch-clustered bootstrap ($p\le 0.003$). Learned anchor-conditioned heads, including a residual head trained on top of the closed-form estimator, do not improve on it. The gain is therefore not evidence that the model ``understands the shoot''; it is evidence that the shoot-level offset is low-dimensional and estimable from a few edits.

Fourth, risk and preference are separate outcomes. A prior trained with unedited negatives already produces a cold-image false-edit rate of 0.99\% (predicted exposure magnitude $\ge 0.3$\ev{} on 302 held-out unedited images). Adding a logistic or MLP gate raises the false-edit rate to 1.99--7.28\% without changing edit recall or test MAE. In an exploratory 50-image blind comparison, the most-preferred arm was the one with the lowest parameter error in only 18.4\% of cases and the lowest rendered error in 14.3\%; preference and fidelity should not be used interchangeably.

\paragraph{Contributions.}
(1) A batch-shared measurement protocol that separates image content, user prior, and current-shoot context, with anchors excluded from targets and inference clustered at the batch level.
(2) A longitudinal, real-Lightroom study of representation choice (absolute vs.\ prior-relative), negative-trained priors, and same-batch adaptation, including the cross-batch negative transfer of the residual representation.
(3) Evidence that a closed-form shrinkage estimator is a strong, auditable baseline for same-batch adaptation, and that learned anchor-conditioned variants do not beat it at this data scale.
(4) An explicit boundary analysis: one photographer, catalog-as-batch proxy, six global parameters in the main study (thirteen in Appendix~\ref{app:e8}), a render-space check against native Lightroom exports, and a clear separation between parameter fidelity, cold-start risk, and preference.

\section{Related Work}

\paragraph{Parameter prediction and white-box editing.}
Exposure-style systems learn a sequence of white-box filters from a set of style photographs~\cite{hu2018}; WISE learns white-box stylization from examples~\cite{loetzsch2022wise}; neural spline transforms model personalized color curves~\cite{bianco2020splinet}; Deep Preset predicts the settings of a low-level color transformation, including Lightroom presets, from a reference pair~\cite{ho2021deepPreset}. Deep bilateral methods learn image-adaptive local operators~\cite{gharbi2017}. These systems establish that interpretable, parameter-space supervision is viable, but they learn a style set rather than a single photographer's longitudinal history, and they do not separate retrieval, prior, and shoot context.

\paragraph{Personalized and retrieval-based editing.}
Adobe's patent describes retrieving stored adjustments for similar images~\cite{adobe2016patent}. Commercial services learn from a user's catalog, and the open-source StyleAI combines image embeddings with a vector store and absolute Develop targets~\cite{styleai}. Recent retrieval-augmented retouching conditions a generative model on reference images without test-time fine-tuning~\cite{weldengus2026refretouch}. PPS learns a user's photographic style from pairwise preference judgments across 767 users~\cite{kim2026pps}; PIF transfers style from a few reference images via textual inversion~\cite{zhu2025pif}; commercial AI profiles train on thousands of a photographer's edits and predict sliders per image (e.g., Imagen AI~\cite{imagen}). These systems personalize \emph{what} the final image should look like. Our question is different: \emph{which part} of a photographer's adjustment is shoot-specific, and how many same-shoot edits are needed to recover it. We therefore treat them as complementary rather than competing: a strong per-image style model can sit on top of the batch-level offset we measure.

\paragraph{User modeling.}
PieNet models a user's taste from comparisons~\cite{kim2020pienet}, and related work embeds users into a shared preference space. These methods answer ``what does this user like?''; we ask ``what part of this user's adjustment is recoverable from the current shoot?'' The two questions are complementary, and our negative results constrain the second.

\paragraph{Presets and shoot-level consistency.}
Presets are the industry's manual answer to shoot-level consistency. Our batch anchors can be read as an automatically estimated, per-shoot preset with a shrinkage weight. Batch editing has also been framed as transferring one demonstrated edit to a set of images~\cite{nguyen2024editone}; we study the statistical version---estimating a shoot-level offset from a few edits---and evaluate it at the batch level. The difference is measurement: we report the full $k$-curve, the batch-macro and worst-batch behavior, and the fact that a few anchors already capture most of the gain.

\paragraph{Evaluation.}
Parameter error, rendered-image error, and human preference are different targets. Recent reward models evaluate instruction-guided edits against human preferences~\cite{wu2026editreward}. Our results add a caution: in a small single-photographer study, the arm preferred by the photographer was frequently not the arm with the lowest parameter or rendered error, so the choice of metric is itself a claim about what personalization means.

\section{Data}

\paragraph{Source and unit.}
The data are a private Lightroom catalog collection from one professional photographer, scanned read-only. The catalogs contain 5,325 photographs with non-default values in the six modeled fields. Of these, 3,196 have an original file on the archive and a reconstructed ``before'' preview; the remaining 2,129 are catalog-only because their originals were not archived. The analysis unit is a photograph, grouped by its directory-level batch, a proxy for a shoot or session.

Each photograph contributes a six-dimensional label read from the Develop settings: exposure (EV), contrast, saturation, vibrance, clarity, and texture. Contrast and saturation are stored as multipliers $1+v/100$; vibrance, clarity, and texture are stored as $v/100$. The six fields are the only global adjustments modeled. Local masks are present in 67.3\% of the photographs and are not predicted.

\paragraph{Splits and exclusions.}
Splits are formed at the batch level, never by random photograph. The 26 directory-level batches are partitioned into 15 training/validation batches and 11 test batches. Training uses 1,571 photographs from those 15 batches and validation 174 photographs drawn from 14 of them; the 11 test batches contain 1,448 photographs in three groups: four recent batches (217 photographs), four unseen batches (790), and three second-group batches (441). Validation is therefore an in-domain validation set, not a batch-level holdout. All reported test numbers use the 11 held-out batches (Table~\ref{tab:splits}).

Three photographs (indices 3123--3125) sit at a batch-definition boundary: their catalog is one of the test batches, but their files live in a directory otherwise assigned to training/validation. Their preview hashes are unique, but their SigLIP similarity to test-batch photographs reaches 0.98, consistent with the same shoot. Because the correct assignment is ambiguous, we exclude them from training and validation; the test set is unchanged. The v1 collection overlaps 1,329 photographs; 1,867 are new. Seventy-four v1 labels changed because the catalog was edited again after the v1 scan; v2 uses the current values. Appendix~\ref{app:v1} reports v1 results separately and they are never pooled with v2.

\paragraph{After images.}
The photographer also keeps the native Lightroom exports: 6,347 JPEGs across 48 export folders, with the original filenames. Matching by filename and a content fingerprint links 2,751 of the 3,196 records (86.1\%) to exactly one export (2,702 in the same folder, 49 across folders after a perceptual check); the remaining 445 were not exported or belong to catalogs whose originals are not archived. Of the 1,448 test photographs, 1,226 (84.7\%) have an export. Section~\ref{sec:render} uses the uncropped subset of these pairs to check the parameter-space ranking in render space.

\begin{table}[t]
\centering
\small
\caption{Post-exclusion splits. Validation is in-domain.}
\label{tab:splits}
\begin{tabular}{lr}
\toprule
Split & $n$ \\
\midrule
Train & 1,571 \\
Validation (in-domain) & 174 \\
Test: recent & 217 \\
Test: unseen & 790 \\
Test: second group & 441 \\
\midrule
Test total & 1,448 \\
Excluded (boundary) & 3 \\
\bottomrule
\end{tabular}
\end{table}

\paragraph{Batch composition and heterogeneity.}
Table~\ref{tab:batches} anonymizes the 11 test batches. Batch sizes range from 38 to 338 photographs; mean exposure varies from $-0.32$ to $+1.31$\ev{}. A one-way decomposition of the full 3,196-label set attributes 36\% of exposure variance and 27\% of contrast variance to batch identity (saturation 10\%, vibrance 13\%, clarity 14\%, texture 15\%), and batch medians relative to the global median range from $+0.84$ to $-0.62$\ev{}. This is descriptive, not causal: catalogs differ in subject, light, camera, and the photographer's intent, and the catalog is only a proxy for the shoot.

\begin{table}[t]
\centering
\small
\caption{Anonymized held-out batches.}
\label{tab:batches}
\begin{tabular}{llrr}
\toprule
Batch & Group & $n$ & Mean exp.\ (EV) \\
\midrule
B01 & recent & 41 & $-0.32$ \\
B02 & recent & 57 & $+0.24$ \\
B03 & recent & 38 & $+1.00$ \\
B04 & recent & 81 & $-0.19$ \\
B05 & unseen & 338 & $+1.15$ \\
B06 & unseen & 246 & $+0.46$ \\
B07 & unseen & 146 & $+1.31$ \\
B08 & unseen & 60 & $+0.32$ \\
B09 & second group & 149 & $-0.14$ \\
B10 & second group & 222 & $-0.17$ \\
B11 & second group & 70 & $+0.36$ \\
\bottomrule
\end{tabular}
\end{table}

\paragraph{What the data do not contain.}
The 2,751 photographs with a native export give real after images, but they are the photographer's final edits, not a neutral reference, and our renderer lacks Adobe's camera profile; Section~\ref{sec:render} therefore uses them for relative comparisons only, on the uncropped subset. Reconstructing after images from catalog parameters was tested and rejected: the preview's auto-bright baseline differs from Lightroom's neutral rendering by 1--3\ev{}, so a reconstructed image is not a valid reference. The paper evaluates parameter fidelity as its primary outcome, uses rendering as a rank check, and treats the 50-image blind study as exploratory.

\section{Method}

\paragraph{Features and prior.}
Each photograph is encoded by a frozen SigLIP-large encoder (1,024 dimensions) concatenated with 84 hand-crafted features, for 1,108 dimensions. The encoder is not fine-tuned. The deployable prior is an MLP ($1{,}108\to192\to9$, about 0.22M parameters) trained on the 1,571 training photographs with AdamW, batch size 64, 60 epochs, dropout 0.1, and model selection on the 174 validation photographs. Nine output slots are retained for compatibility with the production runtime; six are supervised, and white balance, tint, and dehaze are held at neutral.

The prior used throughout is \textbf{V25-neg}, trained with 602 unedited cold images added as negatives at weight 0.3 (a disjoint set of 603 cold images is held out; 301 calibrate gates and 302 test them). The negative term teaches the prior to output near-default adjustments on photographs the photographer did not edit. On the 302 cold test images, the prior's false-edit rate is 0.99\%, defined as the fraction with $|\hat{e}|\ge0.3$\ev{}.

\paragraph{Absolute and residual heads.}
Two heads share the architecture. The \textbf{absolute} head predicts the six labels directly. The \textbf{residual} head predicts $\Delta=y-P(x)$, with its final layer zero-initialized so training starts exactly at the prior; at inference the prediction is $P(x)+\beta\hat{\Delta}$ with $\beta=1$ in the main table. A $\beta$ sweep (0 to 1.5) is flat between 1.0 and 1.5 on validation. Both heads train for 60 epochs with seed 0. The absolute head is trained without negatives; a separate variant adds the 602 cold negatives at weight 0.6. The residual head is trained without negatives, following the earlier finding that the negative term hurt residual targets.

\paragraph{Batch-shared anchor protocol.}
The deployable question is: given $k$ already-edited photographs from the current batch, how well can we predict the remaining photographs? For each test batch and repetition, we sample $k$ anchors without replacement and evaluate on the rest of that batch; all targets share the same anchor set. This is the \emph{batch-shared} protocol. It differs from per-target sampling, where each target draws its own anchor set; that scheme does not correspond to a batch workflow and can shift the estimated curve in either direction, so we treat it only as a diagnostic.

The main curve uses $k\in\{0,1,2,4,8,16,32\}$, 20 repetitions, seed 7. Batches with $n<k+1$ use $k=n-1$; at $k=0$ every estimator degenerates to the prior. The anchor-variants experiment uses the same protocol with seed 21 and $k\in\{4,8,16\}$. No anchor is ever also a target, and no training-batch photograph is used as an anchor for a test batch.

\paragraph{Estimators.}
Let $P_i$ be the prior prediction and $A$ the anchor set.
\emph{Prior:} $P_i$.
\emph{Batch median:} the per-field median of $y_a$ over $a\in A$.
\emph{Residual median:} $P_i+\operatorname{median}_{a\in A}(y_a-P_a)$.
\emph{Hier-EB:} $P_i+w\odot\operatorname{median}_{a\in A}(y_a-P_a)$, with $w_j=k/(k+\kappa_j)$, where $\kappa_j$ is a training-batch noise-to-signal ratio: the mean within-batch variance of $y-P$ divided by the variance of per-batch residual means.
\emph{Mix50:} $0.5\cdot\operatorname{median}_{a\in A}(y_a)+0.5\cdot P_i$.
\emph{Online EB (E6d):} the same shrinkage form, with $\kappa_j=\operatorname{Var}_A(y_j-P_j)/\operatorname{Var}_A(y_j)$ estimated from the current anchors; it requires no historical batch statistics and falls back to the prior at $k=0$.

We also compare three learned anchor-conditioned variants under the same protocol: concat (image features plus a 13-dimensional anchor summary), median+residual, and E6d+residual. They test whether content features add information beyond the offset once the offset is explicitly estimated.

\paragraph{Cold-start gate and metrics.}
Logistic and MLP gates are trained on the 1,571 edited training photographs and the 602 cold negatives; thresholds are set on the 301 cold calibration images to keep a false-positive rate $\alpha\in\{0.05,0.10\}$. A rejected photograph receives the default (no edit). The primary metric is exposure MAE in EV; we also report nMAE6 (per-photograph mean $|$error$|$/span over six fields, then averaged) and cold false-edit rate. For anchors we report the micro average, the batch-macro average over 11 batches, and the worst batch.

Confidence intervals use a \emph{batch-clustered bootstrap}: we resample the 11 test batches with replacement 10,000 times, keeping all photographs within a sampled batch, and recompute the paired difference. This respects correlation among photographs from the same shoot and is the only interval used for main claims. With 11 clusters the intervals are wide; $p$-values are descriptive rather than confirmatory. Photograph-level tests are reported only to show how much they overstate precision.

\begin{figure}[t]
\centering
\begin{tikzpicture}[
  node distance=4mm and 4mm,
  box/.style={draw=gray!60, fill=gray!6, rounded corners=1.5pt,
              align=center, font=\scriptsize, inner sep=3pt, text width=3.15cm},
  arr/.style={-{Latex[length=1.4mm]}, draw=gray!70, line width=0.5pt}
]
\node[box] (hist) {Lightroom history\\41 catalogs, 3,193 photos};
\node[box, right=of hist] (prior) {frozen encoder\\$\rightarrow$ prior $P$};
\node[box, below=of hist] (anch) {current batch\\$k$ edited anchors};
\node[box, right=of anch] (shrink) {closed-form shrinkage\\mix50 / hier-EB / online EB};
\node[box, below=of shrink] (pred) {prediction for the\\remaining photographs};
\draw[arr] (hist) -- (prior);
\draw[arr] (prior) -- (shrink);
\draw[arr] (anch) -- (shrink);
\draw[arr] (shrink) -- (pred);
\end{tikzpicture}
\caption{Signal decomposition. A content prior and a few same-batch anchors are combined by closed-form shrinkage; the evaluation unit is the batch, and anchors are never targets.}
\label{fig:pipeline}
\end{figure}
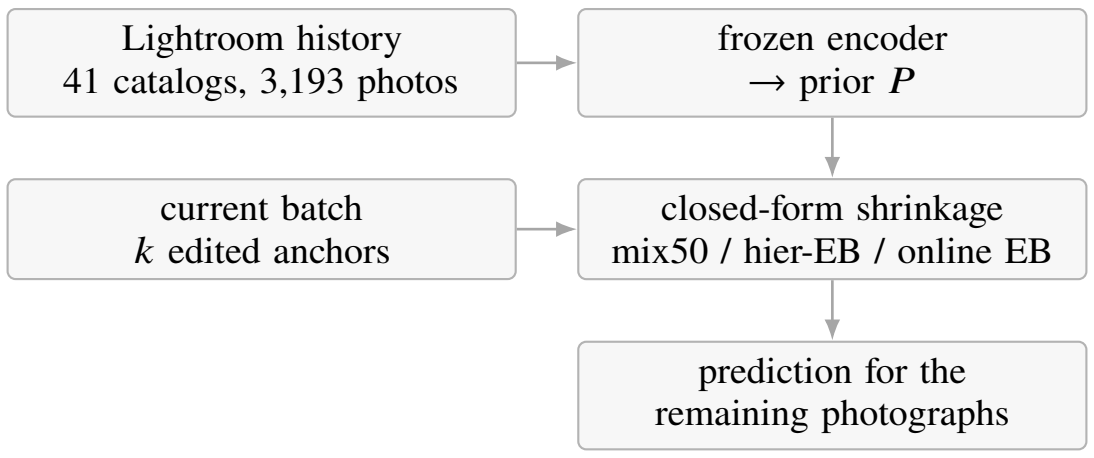

\section{Results}

\subsection{Batch identity is a first-order signal}
The decomposition in Section~3 attributes a substantial share of label variance to batch identity (36\% exposure, 27\% contrast), and batch medians span more than 1.4\ev{}. A predictor evaluated per photograph, with photographs randomly split, would see many near-duplicate labels from the same shoot in both train and test. This motivates both the batch-level splits and the batch-shared anchor protocol, and explains why a content-only model can look better in-domain than it is on a new shoot. The effect is not a camera effect: the body explains 1.2\% of exposure variance against 35.9\% for batch, and 20 of the 26 batches mix both bodies (mean within-batch body $\eta^2=0.09$).

\subsection{Residual targets help in-domain, not across batches}
\begin{table}[t]
\centering
\small
\caption{Absolute vs.\ residual heads (V25-neg prior, post-exclusion).}
\label{tab:e5}
\begin{tabular}{lrrr}
\toprule
Split ($n$) & Prior & Absolute & Residual \\
\midrule
Validation (174) & 0.1811 & 0.1950 & \textbf{0.1715} \\
Test, all (1,448) & 0.5856 & \textbf{0.5824} & 0.5925 \\
Test, recent (217) & 0.4030 & 0.4246 & 0.4024 \\
Test, unseen (790) & 0.7442 & 0.7353 & 0.7614 \\
Test, second group (441) & 0.3913 & 0.3860 & 0.3834 \\
\bottomrule
\end{tabular}
\end{table}

On validation, the residual head reduces exposure MAE by 0.024\ev{} relative to the absolute head and by 0.010\ev{} relative to the prior. On the full test set the ordering reverses: the residual head is 0.010\ev{} worse than the absolute head. The paired, batch-clustered difference (residual $-$ absolute) is $+0.0101$\ev{}, 95\% CI $[-0.0158,+0.0329]$, $p=0.51$; the batch-macro difference is $+0.0043$\ev{}. The photograph-level paired test gives $p=0.0266$, which is what an i.i.d.\ analysis would report; it is not the number we use.

The group breakdown shows where the reversal comes from: the residual head is competitive on the recent and second-group batches (0.402 vs.\ 0.425 and 0.383 vs.\ 0.386) but worse on the unseen batches (0.761 vs.\ 0.735), which contain most of the extreme exposure labels (80 of the 99 photographs with $|\text{exposure}|>2$). The absolute head beats the prior on the two larger test groups (unseen 0.735 vs.\ 0.744; second group 0.386 vs.\ 0.391), although it is worse than the prior on the recent group (0.425 vs.\ 0.403); the residual head is close to the prior on recent batches but clearly worse on unseen. We therefore treat the residual representation as a \emph{descriptive, in-domain} result, not a general improvement. The v1 experiment, with a different split and prior, showed a larger in-domain gain (Appendix~\ref{app:v1}); it does not replicate as a cross-batch claim in v2.

\subsection{A negative-trained prior handles cold start}
The prior alone reaches 0.5856\ev{} on the full test set. Adding the negative term to the absolute head changes test MAE from 0.5824 to 0.5890\ev{}, a small edit-error cost; the deployed negative-trained prior reaches 0.5856\ev{}. The residual head, trained without negatives, reaches 0.5925\ev{}. On the 302 cold test images the prior's false-edit rate is 0.99\%---already low: it rarely proposes a visible exposure change on photographs the photographer left untouched. The trade-off is favorable in the deployed configuration, but it is a trade-off, not a free improvement.

\subsection{Same-batch anchors are the robust signal}
\begin{table}[t]
\centering
\small
\caption{Batch-shared anchor curve (V25-neg, $n=1{,}448$, 20 repetitions, seed 7). Online EB uses seed 21 and $k\in\{4,8,16\}$.}
\label{tab:anchors}
\begin{tabular*}{\columnwidth}{@{\extracolsep{\fill}}rrrrr@{}}
\toprule
$k$ & Prior & Mix50 & Hier-EB & Online EB \\
\midrule
0 & 0.5856 & 0.5856 & 0.5856 & 0.5856 \\
1 & 0.5856 & 0.5186 & 0.5343 & -- \\
2 & 0.5872 & 0.4934 & 0.5252 & -- \\
4 & 0.5863 & \textbf{0.4657} & 0.4739 & 0.4805 \\
8 & 0.5896 & 0.4717 & 0.4650 & \textbf{0.4573} \\
16 & 0.5920 & 0.4712 & \textbf{0.4525} & 0.4559 \\
\bottomrule
\end{tabular*}
\end{table}

\begin{figure}[t]
\centering
\includegraphics[width=\columnwidth]{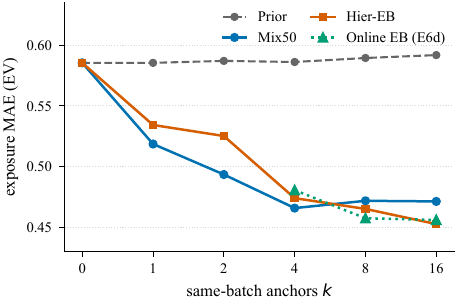}
\caption{Exposure MAE as a function of the number of same-batch anchors. The best closed-form arm changes with $k$; differences among arms at a fixed $k$ are 0.003--0.015\ev{}.}
\label{fig:curve}
\end{figure}

The curve falls steeply from $k=0$ to $k=4$ (Table~\ref{tab:anchors}, Figure~\ref{fig:curve}). Beyond that, hier-EB continues to improve slowly (0.4739 to 0.4525 from $k=4$ to 16), while mix50 is essentially flat (0.4657 to 0.4712). At $k=4$, mix50 is best (0.4657); at $k=8$, online EB is best (0.4573); at $k=16$, hier-EB is best (0.4525). The differences among the three closed-form arms at a fixed $k$ are small (0.003--0.015\ev{}) and the batch-clustered intervals do not separate them; no single arm dominates at all $k$.

Learned anchor-conditioned variants do not beat the closed form (Table~\ref{tab:variants}): concat is close to the prior, median+residual is close to the median, and adding a residual head on top of online EB changes the result by 0.003--0.005\ev{} in the wrong direction. The online EB is also more stable to training-batch composition: after excluding the three boundary photographs, hier-EB at $k=4$ moved from 0.4591 to 0.4739 because the training-batch $\kappa$ changed, while online EB at $k=4$ stayed at 0.4805.

\begin{table*}[t]
\centering
\small
\caption{Anchor variants (V25-neg, $n=1{,}448$, seed 21). Bold marks the best column within this table; hier-EB reaches 0.4525 at $k=16$ in Table~\ref{tab:anchors}.}
\label{tab:variants}
\begin{tabular}{rrrrrrr}
\toprule
$k$ & Prior & Median & Online EB & Concat & Median+resid. & E6d+resid. \\
\midrule
4 & 0.5871 & 0.5517 & \textbf{0.4805} & 0.5807 & 0.5783 & 0.4852 \\
8 & 0.5896 & 0.5180 & \textbf{0.4573} & 0.5863 & 0.5278 & 0.4607 \\
16 & 0.5934 & 0.5059 & \textbf{0.4559} & 0.5903 & 0.5294 & 0.4598 \\
\bottomrule
\end{tabular}
\end{table*}

For hier-EB, the batch-macro average falls to 0.434 and the worst batch to 0.796 at $k=8$; at $k=16$, to 0.422 and 0.760. The improvement is visible at the macro and worst-batch level, not only in the micro average. The worst batch remains the one with the largest exposure shift (1.24\ev{} under the prior), consistent with anchors capturing the direction but not the full magnitude of an extreme shoot.

\begin{table}[t]
\centering
\small
\setlength{\tabcolsep}{3pt}
\caption{Paired differences vs.\ prior, batch-clustered 95\% CIs (11 batches, 10,000 resamples).}
\label{tab:ci}
\begin{tabular*}{\columnwidth}{@{\extracolsep{\fill}}llrrr@{}}
\toprule
Comparison & $k$ & $\Delta$ EV & 95\% CI & $p$ \\
\midrule
Mix50 $-$ prior & 4 & $-0.1200$ & $[-0.221,-0.020]$ & 0.0010 \\
Hier-EB $-$ prior & 4 & $-0.1116$ & $[-0.206,-0.014]$ & 0.0026 \\
Mix50 $-$ prior & 8 & $-0.1171$ & $[-0.208,-0.027]$ & $<0.001$ \\
Hier-EB $-$ prior & 8 & $-0.1229$ & $[-0.228,-0.016]$ & 0.0024 \\
Mix50 $-$ prior & 16 & $-0.1187$ & $[-0.207,-0.028]$ & 0.0002 \\
Hier-EB $-$ prior & 16 & $-0.1353$ & $[-0.252,-0.018]$ & 0.0024 \\
Online EB $-$ prior & 4 & $-0.1057$ & $[-0.211,+0.005]$ & 0.070 \\
Online EB $-$ prior & 8 & $-0.1305$ & $[-0.252,-0.007]$ & 0.0246 \\
Online EB $-$ prior & 16 & $-0.1340$ & $[-0.254,-0.017]$ & 0.0038 \\
\bottomrule
\end{tabular*}
\end{table}

The anchor gain is the only effect that survives batch clustering at every $k\ge4$ for mix50/hier-EB, with an effect of roughly 0.11--0.14\ev{} (Table~\ref{tab:ci}). Online EB is marginal at $k=4$ ($p=0.070$) and significant at $k=8$ and 16; its anchor-estimated $\kappa$ is noisier with very few anchors. Median-only and learned-concat arms are not significant at any $k$ ($p=0.22$--$0.85$).

\begin{figure}[t]
\centering
\includegraphics[width=\columnwidth]{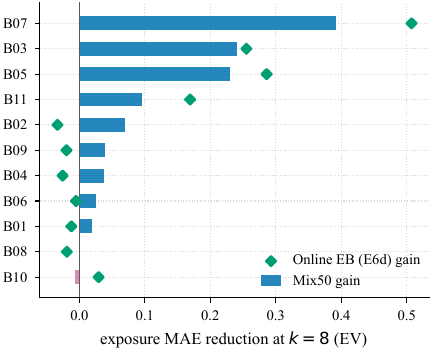}
\caption{Per-batch exposure-MAE reduction at $k=8$, sorted by the mix50 gain. Bars: mix50; diamonds: online EB. The gain is largest in batches with large exposure shifts (B03, B05, B07); one batch (B10) is slightly worse under mix50.}
\label{fig:perbatch}
\end{figure}

Figure~\ref{fig:perbatch} shows that the gain varies by batch. Mix50 is better than hier-EB in 7 of 11 batches, but hier-EB has the lower micro average because its wins fall in the larger batches. This is why we report micro and macro and do not declare a single best arm.

\paragraph{What the anchors fix.}
At $k=8$ the gain is concentrated in exposure: MAE falls from 0.589 to 0.463 (21\%), while contrast, vibrance and clarity improve by 7--9\% and saturation and texture by less than 2\%. Anchors correct the shoot's tonal offset, not its colour or detail treatment.

\paragraph{Which anchors?}
We also ask whether the choice of anchors matters. Fixing the evaluation half of each test batch and drawing anchors from the other half, we compare six rules at $k\in\{2,4,8,16\}$: random; the first $k$ frames of the shoot; the $k$ largest-$|$exposure$|$ edits; the $k$ photographs closest to the batch centroid in feature space; farthest-point sampling in that space; and an oracle that picks the $k$ closest to the batch label median. At $k=4$ exposure MAE is 0.467 for random, 0.671 for first-$k$, 0.563 for extreme edits, 0.550 for the feature medoid, 0.528 for feature diversity, and 0.450 for the oracle (batch-clustered $p=0.032$ against random). Only the oracle's advantage is significant: no deployable content-based rule beats random, and the opening frames are clearly worse. Selection therefore has little deployable headroom---the protocol's random anchors are not leaving accuracy on the table, and a user need not choose which frames to edit.

\subsection{Protocol and prior choice change the answer}
Three choices materially change the reported result. First, \emph{per-target anchor sampling} gives each target its own anchor set; it yields 0.4762/0.4603/0.4503 at $k=4/8/16$, close to and sometimes above the batch-shared arms. Because the protocol differs, we treat it only as a diagnostic. Second, \emph{leave-one-out over photographs} is more optimistic: the same online EB reaches about 0.433\ev{} under LOO, versus 0.457 under batch-shared sampling at $k=8$. Third, \emph{the prior version} matters: the older V20e prior gives 0.469/0.455/0.453, while V25-neg gives 0.4805/0.4573/0.4559. None of these numbers should be compared across priors. The practical rule is narrow: report the batch-shared curve, the batch-macro average, the worst batch, and a batch-clustered interval; do not report a single LOO or per-target number as the deployable result.

\subsection{Under a negative-trained prior, an extra gate is redundant}
\begin{table}[t]
\centering
\small
\caption{Cold-start false-edit rate and gates. Recall and test MAE are on the edited test sets.}
\label{tab:cold}
\begin{tabular}{lrrr}
\toprule
Predictor & Cold FP & Recall & Test MAE \\
\midrule
Prior (V25-neg) & \textbf{0.99\%} & 100\% & 0.5856 \\
LR gate, $\alpha=0.05$ & 1.99\% & 100\% & 0.5856 \\
LR gate, $\alpha=0.10$ & 6.29\% & 100\% & 0.5856 \\
MLP gate, $\alpha=0.05$ & 4.64\% & 100\% & 0.5856 \\
MLP gate, $\alpha=0.10$ & 7.28\% & 100\% & 0.5856 \\
\bottomrule
\end{tabular}
\end{table}

\begin{figure}[t]
\centering
\includegraphics[width=\columnwidth]{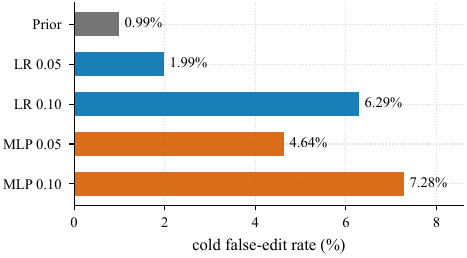}
\caption{Cold false-edit rate. The negative-trained prior is already at 0.99\%; adding a gate increases false edits.}
\label{fig:cold}
\end{figure}

The gates never fire on the edited test photographs at these thresholds (recall 1.0), so they are a no-op on the edit sets and only add false edits on cold images (Table~\ref{tab:cold}, Figure~\ref{fig:cold}). A distance gate is worse: its recall on new batches is 0.1--8\%, so it would suppress most genuine edits. Under the V25-neg prior the extra gate is redundant or harmful. The earlier v1 result---a gate reducing cold false edits from 51\% to 7\%---does not transfer to this prior and split; it is historical, not a claim about the current system.

\subsection{Exploratory: fidelity and preference are distinct}
In an exploratory blind study, 50 photographs spanning the test batches were shown with three anonymized, order-randomized candidates: a production model (v7), a CLIP-based prior (V25-CLIP), and a SigLIP prior with same-batch anchors. The photographer preferred v7 in 44\% of cases, V25-CLIP in 34\%, and SigLIP+anchors in 20\% (one image was rejected by all). However, the preferred arm was also the arm with the lowest absolute exposure error in only 18.4\% of images, and the lowest rendered $\Delta E_{00}$ in 14.3\%; chance is 33\%. Parameter and rendered fidelity therefore did not predict preference in this sample. We report this as a boundary observation, not a preference study: one photographer, a simplified preview without highlight recovery, and no controlled preference protocol. A formal study needs multiple raters and a blinded A/B design; Section~\ref{sec:render} gives the pixel-level check that this exploratory study cannot.

\subsection{Render-space validation against native exports}
\label{sec:render}

Parameter error is not pixel error. For test photographs with a native export we render the same before image under each arm and compare it with the photographer's output. Of the 1,226 paired test photographs, 777 have an export with the original frame's aspect ratio; cropped exports are excluded because we cannot reproduce the crop. The before image is decoded from the RAW with fixed settings (camera white balance, no auto-brightness), rendered at 512\,px under each arm, and compared with the export using $\Delta E_{00}$ and SSIM.

Two conditions make the comparison informative. The renderer must follow Lightroom's exposure and contrast semantics, so we calibrate it on separate export ladders (linear-light $2^{EV}$ exposure; a soft-pivot contrast S-curve fitted on three scenes at nine settings, RMS 0.001--0.018 versus 0.04--0.23 for a straight stretch). Its absolute baseline also differs from Adobe's by a per-batch exposure offset of $-0.5$ to $+2.0$\ev{}, which we calibrate once per batch on a disjoint half (near-neutral edits only), evaluating on the other half: 383 photographs.

\begin{table}[t]
\centering
\footnotesize
\setlength{\tabcolsep}{3pt}
\caption{Render-space check against 383 native Lightroom exports. Lower $\Delta E_{00}$ and higher SSIM are better. Differences are paired within photograph and clustered by batch (10,000 resamples); intervals are 95\%.}
\label{tab:render}
\begin{tabular}{lcccc}
\toprule
Arm & $\Delta E_{00}$ & vs.\ prior & SSIM & vs.\ prior \\
\midrule
No edit & 13.91 & $-0.03$ & 0.5104 & --- \\
Prior & 13.94 & --- & 0.5033 & --- \\
mix50 ($k{=}8$) & 13.70 & $-0.24$ [$-0.87$, $+0.08$] & 0.5149 & $+0.012$ \\
Online EB ($k{=}8$) & 13.53 & $-0.41$ [$-1.03$, $-0.02$] & 0.5241 & $+0.021$ \\
True parameters & 13.16 & $-0.78$ [$-1.94$, $-0.08$] & 0.5342 & $+0.031$ \\
\bottomrule
\end{tabular}
\end{table}

Table~\ref{tab:render} reproduces the parameter-space ordering: true parameters $<$ online EB $<$ mix50 $<$ prior $\approx$ no edit. At $k=8$, online EB lowers $\Delta E_{00}$ by 0.41 [0.02, 1.03] and raises SSIM by 0.021 against the prior ($p=0.042$ and $p=0.006$), while the true parameters lower $\Delta E_{00}$ by 0.78 [0.08, 1.94]. The effect sizes are small: the absolute $\Delta E_{00}$ of about 13 is dominated by the difference between our renderer and Adobe's camera profile, so only relative comparisons are meaningful. Within that limit, the shoot-level gain measured in parameter space does move the rendered image toward the photographer's export. The same pattern holds when the action space is extended beyond six sliders: a same-batch median look moves the render toward the export, while a per-photograph residual head does not (Appendix~\ref{app:e8}).

\section{Discussion}

\paragraph{Where personalization resides.}
In this dataset, the recoverable signal is overwhelmingly the shoot-level offset. A few same-batch edits reduce exposure error by 0.11--0.14\ev{}, and the reduction survives clustering by batch. Image content, as captured by a frozen general-purpose encoder and a small head, adds little once the offset is estimated: learned anchor-conditioned heads do not beat closed-form shrinkage, and the residual representation's in-domain gain does not transfer. This does not mean content is irrelevant to editing; it means that at this data scale and with six global parameters, the content residual is small relative to the batch offset and does not generalize across shoots.

\paragraph{Why the residual target fails across batches.}
The residual head is trained to predict $y-P(x)$. In-domain, the prior and the residual are estimated on the same batches, so the residual absorbs a stable batch-specific offset. On a new batch that offset is different; the residual head has learned to reproduce the training batches' offsets and has no mechanism to update them. The anchor estimators succeed precisely because they estimate the offset from the target batch at inference time. The failure is not an optimization failure; it is a mismatch between the representation and the evaluation unit.

\paragraph{An external stress test.}
On FiveK (four experts, 231 images each; pseudo-labels obtained by inverting rendered outputs), expert identity explains only 1--8\% of target variance in our re-analysis, a mean prior or ridge is as good as memory or a small head, and cross-user models match self-models~\cite{bychkovsky2011}. This is the opposite regime: low style dispersion. It supports the interpretation that batch anchors pay off when shoot-level dispersion is high, and it warns that FiveK is not a personalization benchmark. It also shows that the estimator is not universally beneficial; the gain is conditional on the data regime.

\paragraph{A deployable artifact, not a new architecture.}
The estimator we recommend is closed-form, runs on CPU, and requires no trained weights beyond the prior; it is already wired into a production editor. At $k=0$ the negative-trained prior gives a 0.99\% cold false-edit rate; at $k\ge4$ the anchor estimator reduces exposure MAE by 0.11--0.14\ev{} with batch-clustered intervals excluding zero. This is a usable model even though it is not a new neural network, and its auditability is a feature for a tool that changes a photographer's files.

\paragraph{A practical recipe.}
At $k=0$ use the negative-trained prior; at $k=1$--2 a 50/50 mix of prior and batch median is stable; at $k\ge4$ online EB or hier-EB captures most of the gain, and the choice among closed-form arms should follow batch-macro and worst-batch behavior rather than a single micro average. Do not add a per-photograph gate to a negative-trained prior, and do not use LOO or per-target sampling for deployable numbers.

\paragraph{Implications for ``style learning'' products.}
A product that claims to learn a photographer's style makes two claims: a long-run prior and a per-shoot adaptation. In this workflow the second is the easier and more valuable one, and a closed-form estimator with a handful of anchors delivers it. The first is real but small: the prior beats a trivial baseline, yet its cross-batch error stays high and a content-only model does not close the gap. Vendors should say which signal their evaluation measures.

\section{Limitations}

\begin{enumerate}
\setlength{\itemsep}{1pt}
\item \textbf{One photographer.} All real data come from a single professional photographer. The paper is a longitudinal case study, not a cross-user study. FiveK-style multi-expert data cannot substitute: its parameters are inferred from rendered outputs~\cite{bychkovsky2011}, and in our re-analysis its expert styles are highly convergent.
\item \textbf{Catalog as batch proxy.} Catalogs and directories approximate shoots but do not exactly match capture sessions. Three photographs had to be excluded because catalog and directory disagreed. The batch effect includes subject, light, camera, and editing-session effects.
\item \textbf{Eleven test batches.} The batch-clustered bootstrap has only 11 clusters. Intervals are wide, and $p$-values should be read descriptively. More shoots are the main requirement for stronger inference.
\item \textbf{Six global parameters in the main study.} Appendix~\ref{app:e8} extends the action space to thirteen global controls (highlights, shadows, whites, blacks, eight-band HSL saturation, dehaze) and shows that a same-batch median look transfers, but local masks (67.3\% of catalogs; 75.4\% of paired exports), tone curves, white balance, and the full rendering pipeline remain outside. Conclusions apply to global slider prediction, not complete retouching.
\item \textbf{Render-space evaluation is relative.} Native exports exist for 84.7\% of the test set, and after crop filtering the render check uses 383 photographs. Its absolute $\Delta E_{00}$ (about 13) is dominated by the missing Adobe camera profile and by a per-batch baseline that must be calibrated on a disjoint subset; only the ordering is interpretable. The preference study remains exploratory and single-rater.
\item \textbf{Selection and drift.} The 2,129 catalog-only photographs are missing because their originals were not archived; if they differ systematically, the dataset is biased. Seventy-four v1 labels changed after the v1 scan; v2 uses current values.
\item \textbf{No formal preference or adoption study.} The paper does not claim adoption rates, user satisfaction, or aesthetic quality.
\end{enumerate}

\paragraph{What would change the conclusion.}
A second photographer with low shoot-level dispersion (like FiveK) would shrink the anchor gain; a multi-rater preference study could show that parameter fidelity and preference diverge further; and a different batch definition could reduce the measured batch effect. The claims are therefore conditional on the data regime, not universal.

\section{Reproducibility and Privacy}

We will release the evaluation code, the estimator implementations, the data schema, the split-construction rules, and the SHA-256 hashes of the frozen manifest and split files. We will not release photographs, Lightroom catalogs or sidecars, real slider parameters, image embeddings, batch names, or checkpoints that could memorize training images. Aggregate statistics sufficient to check the main tables are released with the package, together with a synthetic example that exercises the schema. The private dataset can be re-created only by its owner; the released code is designed so that the estimators and the protocol can be run on any compatible parameter-editing dataset. The frozen snapshot hashes (first 16 hex characters) are: manifest \texttt{5b57ec3d8abaff5b}, splits \texttt{b32ae1f9cc42922c}, SigLIP features \texttt{42f64327c560626a}, CLIP features \texttt{7ba52678e4f1fa70}, stage labels \texttt{33a8c2ee73b6edff}. The three excluded photographs are recorded in the split metadata; the pre-exclusion split is retained for audit.

\section{Conclusion}

Personalization in photo editing is often discussed as a single capability. In one photographer's Lightroom history, it decomposes into a weak cross-shoot content signal, a useful long-run prior, and a strong shoot-level offset that can be estimated from a handful of same-batch edits. The shoot-level offset is the part that transfers into a deployable workflow today; the residual representation and the learned anchor-conditioned heads do not. We hope the batch-shared protocol and the negative results here make future personalization claims easier to compare, and we encourage reporting parameter fidelity, cold-start risk, and preference as separate outcomes.

\appendix
\small
\section{Historical v1 results}
\label{app:v1}

The v1 collection contains 1,387 edited photographs, a different split, and the V20e/V24e prior. On v1, the residual head reached 0.1260\ev{} versus 0.1404\ev{} for the absolute head, and a 50/50 residual fusion reached 0.1164\ev{} versus 0.1357\ev{} for the production model. These numbers motivated the v2 study but are not comparable to the v2 tables: the sample, split, prior, and evaluation protocol differ. The v1 gate result (cold false-edit rate reduced from 51\% to 7\%) is likewise historical; under the V25-neg prior the prior alone reaches 0.99\%.

\section{Per-batch behavior at \texorpdfstring{$k=8$}{k=8}}
\label{app:perbatch}

\begin{table}[h]
\centering
\small
\setlength{\tabcolsep}{3pt}
\caption{Per-batch exposure MAE at $k=8$. The online-EB column uses a separate sampling seed.}
\label{tab:perbatch}
\begin{tabular*}{\columnwidth}{@{\extracolsep{\fill}}llrrrr@{}}
\toprule
Batch & Group & $n$ & Prior & Mix50 & E6d \\
\midrule
B01 & recent & 41 & 0.280 & 0.260 & 0.292 \\
B02 & recent & 57 & 0.330 & 0.261 & 0.364 \\
B03 & recent & 38 & 0.850 & 0.608 & 0.594 \\
B04 & recent & 81 & 0.308 & 0.270 & 0.333 \\
B05 & unseen & 338 & 0.868 & 0.637 & 0.582 \\
B06 & unseen & 246 & 0.392 & 0.367 & 0.398 \\
B07 & unseen & 146 & 1.241 & 0.848 & 0.733 \\
B08 & unseen & 60 & 0.290 & 0.291 & 0.310 \\
B09 & second & 149 & 0.415 & 0.376 & 0.434 \\
B10 & second & 222 & 0.332 & 0.338 & 0.303 \\
B11 & second & 70 & 0.532 & 0.437 & 0.363 \\
\bottomrule
\end{tabular*}
\end{table}

Mix50 is better than hier-EB in 7 of 11 batches, but hier-EB has the lower micro average because its wins fall in the larger batches. The arm ranking is therefore batch-dependent; the paper reports micro, macro, and worst-batch numbers and does not declare a single best arm.

\section{Negative results outside the main line}
\label{app:negative}

The following results are documented in the technical report and are not part of the paper's claims: (i) \emph{stage-wise editing}: human workflows average 4.9 stages per photograph, but per-stage prediction with content residuals is worse than the same-batch anchor median at every stage (exposure $0.463\to0.515$, tone $9.93\to13.53$, HSL $4.49\to7.11$); (ii) \emph{render look-back}: re-encoding the image after an oracle exposure correction does not improve later-stage prediction (tone $12.52\to12.74$; HSL $7.55\to7.75$); (iii) \emph{uncertainty heads}: photograph-level uncertainty trained with out-of-fold or leave-one-batch-out targets reaches AUC 0.30--0.48 for $|\text{exposure}|>2$, below the training-free anchor-dispersion signal (0.878--0.892); (iv) \emph{synthetic corruption pretraining}: ATune-style corruption pretraining~\cite{goyal2026atune} degrades test MAE from 0.586 to 0.607/0.633; (v) \emph{anchor-conditioned learning}: concat, median+residual, and E6d+residual are at or above the closed-form online EB; (vi) \emph{crop ranking}: a 4B vision-language model agrees with the photographer in 36.7\% of 60 images, below the constant center-85\% baseline of 40\%; (vii) \emph{quantization}: a 4B model at NF4 fits in 8\,GB (2.86\,GiB peak) but shifts behavior relative to fp16 and is usable only as a suggestion.

\section{Extending the action space: a batch-level full-grade look}
\label{app:e8}

The main study predicts six global adjustments. The catalogs contain more. Across all 5,357 edited photographs in the 41 catalogs (including 2,129 whose originals were not archived), 99.5\% have a non-default highlight, shadow, white, or black value, 93.6\% have a non-identity tone curve, and 57.3\% have an HSL band edit. On the 2,751 photographs with a native export, 36.4\% have a mean band-saturation change below $-10$, and in two thirds of those the global saturation label is within $\pm2$; the correlation between the global saturation label and the mean band saturation is $-0.003$. The six-field target is therefore blind to a large part of the colour work, and the after image is not a clean label for those dimensions.

We therefore built a 13-dimensional extension: highlights, shadows, whites, blacks, eight HSL saturation bands, and dehaze (all in slider units, divided by 100). The protocol is unchanged: 3,196 manifest photographs, batch-level splits (train 1,571, validation 174, test 1,448), and the same 1108-dimensional image features. Table~\ref{tab:e8param} reports parameter MAE. A batch median over eight same-batch anchors reduces MAE from 0.1605 (zero) and 0.0916 (global median) to 0.0823; sixteen anchors reach 0.0789. A per-photograph residual head does not transfer: it reaches 0.0615 on validation but 0.0960 on test (direct prediction 0.1029), and a regularized linear residual is 0.0870--0.0973, all worse than the batch median. The paired degradation against the batch median is $+0.0137$ [0.0094, 0.0195] under a batch-clustered bootstrap. This is the same pattern as the six-field residual head: in-domain validation is misleading, and only the batch-level statistic transfers.

\begin{table}[h]
\centering
\footnotesize
\setlength{\tabcolsep}{3pt}
\caption{Parameter MAE on the 13-dimensional extension (slider units, divided by 100). Lower is better.}
\label{tab:e8param}
\begin{tabular}{lr}
\toprule
Estimator & MAE \\
\midrule
Zero & 0.1605 \\
Global median & 0.0916 \\
Batch median (leave-one-out) & 0.0765 \\
Batch median, $k=8$ & 0.0823 \\
Batch median, $k=16$ & 0.0789 \\
MLP, direct & 0.1029 \\
MLP, $k=8$ + residual & 0.0960 \\
Ridge, $k=8$ + residual & 0.0870--0.0973 \\
\bottomrule
\end{tabular}
\end{table}

\begin{table}[h]
\centering
\footnotesize
\setlength{\tabcolsep}{3pt}
\caption{Render check on 140 paired test photographs (mean over photographs). $\Delta E_{00}$ is against the native export; the $u'v'$ ratio is exposure-invariant and 1.0 is a perfect colour match.}
\label{tab:e8render}
\begin{tabular}{lcc}
\toprule
Arm & $\Delta E_{00}$ & $u'v'$ ratio \\
\midrule
True six parameters & 17.80 & 1.555 \\
+ batch look ($k=8$) & \textbf{16.86} & \textbf{1.236} \\
+ true extended parameters & 16.92 & 1.144 \\
\bottomrule
\end{tabular}
\end{table}

Table~\ref{tab:e8render} renders 140 test photographs against their native exports. We isolate the extension by using the true six parameters in every arm and applying the extended grade with a per-channel 256-point tone LUT; the baseline is the true six parameters alone. Adding the batch look lowers mean $\Delta E_{00}$ from 17.80 to 16.86 and the exposure-invariant $u'v'$ chroma ratio from 1.555 to 1.236 (105/140 photographs improve); the photograph's true extended parameters reach 16.92 and 1.144. Eight anchors therefore capture essentially the whole global gain. The remaining gap is local masks (75.4\% of the paired photographs), the tone curve, and the Adobe camera profile.

Two further controls are negative. A batch-median tone curve has a clear parameter signal (9-point MAE 0.0139 to 0.0065, $-53\%$) but no render gain: adding it to the batch look changes mean $\Delta E_{00}$ by $-0.004$, and the true curve does not help either. The curve operator itself is faithful---a per-channel curve reproduces Lightroom's contrast ladder at 0.54--1.05 $\Delta E_{76}$, versus 1.45--2.77 for a luma-only application---so the null is redundancy with the Basic tone path, not a bad spline. Colour grading has no batch signal (MAE 0.0350 vs.\ 0.0349). Adding batch-median HSL hue and luminance also hurts ($+0.30$ $\Delta E_{00}$, 33/140 improve), so the deployed look keeps saturation only.

The product implementation is closed-form and auditable: a project-look command takes the median of the four tone fields, eight saturation bands, and dehaze over the edited photographs in the same folder (app sidecars or Lightroom XMP, read-only), and adds it to the selected photographs without touching geometry, masks, or the six-field auto-tone snapshot. Tone curves and colour grading are deliberately excluded.

\section{Number provenance}
\label{app:provenance}

All main-table numbers come from the post-exclusion reports of the frozen v2 snapshot. The released package's \texttt{aggregates/main\_results.json} contains the same values; \texttt{dataset\_summary.json} contains the counts and label extremes; \texttt{hashes.json} contains the fingerprints. The mapping is: E5 from the E5-v2 report and its batch-clustered CI file; E6b from the batch-shared anchor curve and its cluster-CI file; anchor variants from the anchor-variants evaluation and its cluster-CI file; E7 from the E7-v2 report; the blind study from the 50-image review log. The appendix's extension numbers come from a separate full-settings scan (\texttt{grade\_full\_2026\_11\_21.jsonl}) and the E8/E8b/E8c/E8d reports; they are an extension of the six-field ledger and are never pooled with the main tables. No number in the paper is entered independently of these files, and the paper and the aggregate file are updated together.

\small
\setlength{\itemsep}{0pt}
\setlength{\parsep}{0pt}

\balance
\end{document}